\documentclass{article} 
\usepackage[preprint]{neurips_2026}

\usepackage[utf8]{inputenc} 
\usepackage[T1]{fontenc}    
\usepackage{microtype}      
\usepackage{xcolor}         

\usepackage{amsmath,amsfonts,bm}

\def\eqref#1{equation~\ref{#1}}

\def\1{\bm{1}}

\DeclareMathAlphabet{\mathsfit}{\encodingdefault}{\sfdefault}{m}{sl}
\SetMathAlphabet{\mathsfit}{bold}{\encodingdefault}{\sfdefault}{bx}{n}

\usepackage{hyperref}
\usepackage{url}
\usepackage{graphicx}
\usepackage{todonotes}
\newcommand{\fitwidth}[1]{\resizebox{\ifdim\width>\linewidth\linewidth\else\width\fi}{!}{#1}}
\usepackage{listings}
\usepackage{booktabs}
\usepackage{algorithm}
\usepackage{algpseudocode}

\title{ZENO: Nonlinear filtering with Zakai \\ Equation Neural Operators}
\author{Aaditya Prasad \\
Picower Institute for Learning \& Memory \\
Department of Brain \& Cognitive Sciences \\
Massachusetts Institute of Technology\\
Cambridge, MA 02139, USA \\
\texttt{aaprasad@mit.edu} \\
\And
Steven W. Flavell \\
Howard Hughes Medical Institute \\
Picower Institute for Learning \& Memory \\
Department of Brain \& Cognitive Sciences \\
Massachusetts Institute of Technology\\
Cambridge, MA 02139, USA \\
\texttt{flavell@mit.edu}
\AND
Brian DePasquale \\
Department of Biomedical Engineering \\
Boston University \\
Boston, MA 02215, USA \\
\texttt{bddepasq@bu.edu} \\
}

\begin{document}

\maketitle

\begin{abstract}
State inference and parameter learning for nonlinear filters is a core computation for scientific machine learning with varied applications in engineering and neuroscience. The unnormalized filtering distribution is governed by the Zakai equation, a stochastic partial differential equation (SPDE) that combines the forward Kolmogorov operator (known as the Fokker-Planck equation) with observed likelihood updates. Although solving these high-dimensional SPDEs is often impossible analytically and intractable numerically, advances in operator learning and physics-informed machine learning have made it feasible. Harnessing these advances, we introduce the \textbf{Z}akai \textbf{E}quation \textbf{N}eural \textbf{O}perator (ZENO), a model that optimizes a neural operator against a Zakai Equation based physics-informed loss to approximate the exact unnormalized filtering distribution for latent stochastic dynamical systems. We use this ZENO-inferred posterior distribution in an expectation-maximization (EM) framework to perform parameter learning and state inference. We demonstrate ZENO's ability to accurately infer the posterior distribution and learn its corresponding SDE parameters in three simulated stochastic dynamical systems (Double-Well bistable system, Van Der Pol (VDP) oscillator, and Lorenz system) and compare its performance against established methods for latent SDE inference and learning. We show that ZENO recovers previously observed limit cycles representing forward and backward locomotion from a whole-brain imaging dataset of immobilized \textit{C. elegans} nematodes. ZENO is a novel approach to the problem of nonlinear filtering and dynamical inference through the lens of partial differential equations and operator theory.
\end{abstract}

\section{Introduction}
Estimating a latent dynamical process from a set of noisy observations is a fundamental problem in machine learning for science \citep{durstewitz_reconstructing_2023}. For example, in neuroscience, it is thought that neural computations are implemented by a latent dynamical system that is expressed by a neural population \citep{saxena_towards_2019}. This problem has been well studied in signal-processing as filtering (in the online setting) or smoothing (in the offline setting) a noisy non-linear dynamical system \citep{yau_filtering_2024}. The essence of filtering as Bayesian inference is to estimate the filtering distribution $p(\mathbf{z}_t | X_{[{0:t}]})$ where $\mathbf{z}_t \in \mathbb{R}^n$ is an unobserved latent variable and $\mathbf{x}_t \in \mathbb{R}^m$ ($m \geq n$) are observations at each point $t \in [0,T]$, while for smoothing the goal is to estimate the posterior distribution $p(\mathbf{z}_t | X_{[0:T]})$ \citep{sarkka_bayesian_nodate}. These distributions are referred to as the filtering and smoothing distributions respectively.

For continuous time systems, the dynamics of the filtering distribution has been well studied as the Kushner-Stratonovich (KS) equation \citep{Stratonovich1960CONDITIONALMP,kushner_64}, a stochastic partial differential equation (SPDE). A more tractable analog of the KS equation is the Zakai equation \citep{zakai_optimal_1969}, a linear SPDE governing the unnormalized posterior density. While solving the KS or Zakai equation provides the exact posterior distribution (post-normalization for the Zakai equation), solving these equations is often impossible analytically and intractable numerically for high-dimensional systems \citep{sabbar2025selectivereviewmodernstochastic}. Alternative methods such as state-space models \citep{bishop_notitle_2006}, (switching) linear dynamical systems \citep{linderman_bayesian_2017}, and deep learning-based latent stochastic differential equations (SDE) \citep{pmlr-v108-li20i} offer a more tractable approach to inference and parameter learning by making assumptions or approximations about the posterior distribution. For example, the Kalman-Bucy filter \citep{kalman_new_1961} provides the exact solution to the KS equation under the assumption that dynamics are linear and observation noise is Gaussian. Thus, the need remains for powerful methods that can approximate the filtering posterior without restrictive assumptions while remaining computationally tractable.

Two recent approaches for solving PDEs with neural networks offer a route around these barriers. \textbf{Physics-informed neural networks (PINNs)} supervise the fit with the governing equation rather than labeled solutions, constraining the network's autodifferentiated derivatives to satisfy the equation at a set of collocation points \citep{Raissi_Perdikaris_Karniadakis_2019,karniadakis_physics-informed_2021}. PINNS are applicable when the solution is never observed, as in filtering, and because no parametric form is imposed, they can represent non-Gaussian
posteriors governing nonlinear dynamics. 

However, PINNs are fit to a single instance of a PDE and must be retrained for each new one, whereas in filtering every observation path $X_{[0,t]}$ drives a distinct instance of the Zakai equation. \textbf{Neural operators} learn the solution operator itself, mapping a PDE's defining data to its solution and generalizing across instances without retraining \citep{lu_learning_2021,azizzadenesheli_neural_2024}, but are trained on labeled solutions which filtering never provides. \textbf{Physics-informed neural operators (PINOs)} resolve both obstacles, using the equation's residual when observations are unavailable to supervise operator learning across a distribution of instances \citep{wang2021learning,li_pino_2024}.

This is the paradigm we build on to solve the Zakai equation for nonlinear filtering. We present the \textbf{Z}akai \textbf{E}quation \textbf{N}eural \textbf{O}perator (ZENO): A physics-informed neural operator for nonlinear filtering that learns the solution to the Zakai equation. ZENO represents the unnormalized filtering distribution by optimizing a physics informed loss which minimizes the residual between the PINO's posterior derivatives and the Fokker Planck operator while incorporating likelihood updates at observation times. Unlike current approaches, which use numerical solvers \citep{leng2026deepzakaijstructuredfiltering} for forecasting or a deep-splitting approach \citep{Beck_2021, bagmark_energy-based_2023,B_gmark_2023,beck2026deeplearningbasednumerical}, we leverage neural operators to learn a global solution over the entire observation history. Furthermore, we make no assumptions about the form of the dynamics model and instead estimate the dynamical parameters and observation model in an expectation-maximization \citep{dempster_maximum_1977} like setup. 

We show that ZENO is able to achieve similar latent inference accuracy to optimal nonlinear filters in simulated dynamical systems. We then show that ZENO performs dynamical parameter estimation at the same level as other state-of-the-art (SOTA) baselines. Finally, we evaluate our model on a real world neuroscience dataset and find that ZENO is able to recapitulate previously observed latent dynamical structure.

\section{Methods}
\subsection{Background}
\subsubsection{Problem Formulation}
The problem of nonlinear filtering involves estimating the posterior distribution $p(\mathbf{z}_t \mid X_{[0,t]})$, where $\mathbf{z}_t \in \mathbb{R}^n$ is an unobserved latent state and $X_{[0,t]} = \{\mathbf{x}_{t_k} : t_k \le t\}$ are noisy $m$-dimensional observations of $\mathbf{z}$ taken at discrete times $t_k = k\Delta t$. The latent state evolves as a stochastic differential equation
\begin{equation}
d\mathbf{z}_t = f(\mathbf{z}_t, t)dt + g(\mathbf{z}_t, t)d\mathbf{W}_t
\label{eq:state_sde}
\end{equation} where $f:\mathbb{R}^n \rightarrow \mathbb{R}^n$ is the deterministic drift function, $g(\mathbf{z},t)$ is an $n \times k$ diffusion coefficient, and $\mathbf{W}_t$ is a $k$ dimensional standard Wiener process. The latent state is observed through a noisy readouts:
\begin{equation}
d\mathbf{x}_t = h(\mathbf{z}_t, t)dt + d\mathbf{V}_t\\
\label{eq:obs_sde}
\end{equation} in the continuous case or
\begin{equation}
    \mathbf{x}_{t_k} = h(\mathbf{z}_{t_k}) + \varepsilon_k, \quad \varepsilon_k \sim \mathcal{N}(0, \mathbf{R})
    \label{eq:obs_discrete}
\end{equation} in the discrete case where $h: \mathbb{R}^n \rightarrow \mathbb{R}^m$ is the observation readout function, $\mathbf{V}_t$ is an $m$-dimensional independent standard Wiener processes, and $\mathbf{R}$ is the observation noise respectively \cite{yau_filtering_2024}. 

\subsubsection{PDEs and SPDEs: Fokker-Planck, Kushner-Stratonovich, and Zakai Equations}
The \textbf{Fokker-Planck} equation (FPE) describes the time evolution of the state's unconditional density, $p(\mathbf{z},t)$, in the absence of any observations, under the dynamics of Equation~\ref{eq:state_sde}:
\begin{equation}
    \frac{\partial p}{\partial t} = \mathcal{L}^*p, \qquad \mathcal{L}^*p = -\nabla \cdot (fp) + \nabla \nabla:(Dp),
\end{equation}
where $D = \frac{1}{2} g(\mathbf{z}, t)g^\top(\mathbf{z},t)$ is the diffusion tensor and $\mathcal{L}^*$ is the forward Kolmogorov operator. The term $\nabla\nabla:(Dp) = \sum_{i,j}\partial_i\partial_j(D_{ij}p)$ is a generalized (anisotropic) Laplacian, reducing to the familiar $D\nabla^2 p$ when $g$ (and hence $D$) is scalar. 

When one only has access to corrupted observations of the state, the conditional PDF of the state $p(\mathbf{z}_t | X_{[0,t]})$ is governed by the \textbf{Kushner Stratonovich} stochastic PDE (SPDE): 
\begin{equation}
    dp = \mathcal{L}^* p dt + p\big(h(\mathbf{z}, t) - \hat{h}(\mathbf{z},t)\big)^\top d\mathbf{I}_t
\end{equation} where $\hat{h}$ is the conditional expectation $\hat{h}(\mathbf{z},t) = \mathbb{E}[h(\mathbf{z},t) | X_{[0,t]}]$ and $dI_t$ is the innovation process, $dI_t = dx_t - \hat{h}(z,t) dt$. For continuous observations (Eq. \ref{eq:obs_sde}), analyzing instead the \textit{unnormalized} probability density function $\rho(\mathbf{z}_t\mid X_{[0,t]})$ yields a more tractable analog of the Kushner-Stratonovich equation — the linear \textbf{Zakai} equation:
\begin{equation}
    d\rho = \mathcal{L}^*\rho dt + \rho h(\mathbf{z}, t) d\mathbf{x}_t .
\end{equation} 

With the discrete observations (Eq. \ref{eq:obs_discrete}), the Zakai equation reduces to the Fokker--Planck equation between observation times and to a multiplicative likelihood update at each observation, giving the recursion
\begin{equation}
    \rho(\mathbf{z}_t|X_{[0,t-\Delta t]}) =(e^{\Delta t \mathcal{L}^*}\rho(\cdot|X_{[0,t-\Delta t]}))(\mathbf{z})
\end{equation}
\begin{equation}
    \rho(\mathbf{z}_t|X_{[0,t]}) = p(\mathbf{x}_t|\mathbf{z}_t)\rho(\mathbf{z}_t|X_{[0,t-\Delta t]})
\end{equation}

Solving the Zakai equation provides the unnormalized filtering distribution but in practice requires numerical methods such as finite-differences which scale poorly due to their grid dependency.

\begin{figure}[h!]
    \centering
    \includegraphics{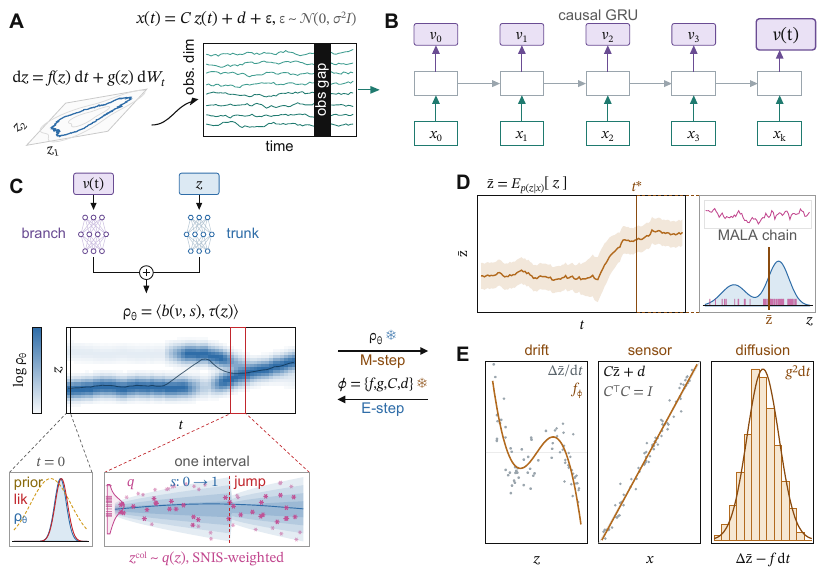}
    \caption{\textbf{Architecture Diagram.} \textbf{A)} Generative Model. \textbf{B)} Causal GRU encodes observations into context trajectory. \textbf{C)} Context trajectory is fed into the branch of a Deep Operator Network along with substeps to produce basis functions. Trunk network can be queried at any $\mathbf{z}$ and outputs $\log \rho_{\theta}$, the unnormalized log density. IC loss between the learned posterior and an IC prior $\mathcal{N}(0,\mathbf{I})$ combined with Fokker Planck Residual Loss between observations and Jump Loss at observations \textbf{D)} Posterior Expectation $\bar{z} = \mathbb{E}_{p(z|x)}[z]$ computed through MALA based MCMC sampling. \textbf{E)} Dynamics and Observation Parameters fit by regressing against $\bar{\mathbf{z}}$}
    \label{fig:1}
\end{figure}

\subsubsection{Physics Informed Neural Operators (PINOs)}
In functional analysis, an operator is a function-to-function mapping. Differential equations are naturally expressed in terms of operators: the equation itself defines one operator acting on the unknown function, and its solution — when it exists — defines another, the operator that maps the equation's data to that solution. For example, for the Fokker-Planck equation, the operator of interest is $\mathcal{G}: p_0 \mapsto p(\cdot,t)$, mapping an initial density to the solution at later times. 

Neural operators are the generalization of neural networks from functions between finite dimensional vector spaces to infinite dimensional functional spaces (Appendix~\ref{app:operator_primer}). They are commonly used to learn the solution operators to differential equations, especially PDEs. In our setting, we seek to learn an operator over the Zakai equation, $\mathcal{G}_\theta: X_{[0,t]} \mapsto \rho_\theta(\cdot,t)$, mapping an observation path directly to the corresponding estimated unnormalized filtering density. 

While neural operators are data-driven, \textbf{Physics Informed Neural Operators} (PINOs) extend them by incorporating the governing differential equation of the system whose solution is being learned (Appendix ~\ref{app:PINN_primer}). This is enforced by ensuring
\begin{equation}
    \label{eq:PINO}
    \frac{\partial \rho_\theta}{\partial t} \approx \mathcal{L}^*\rho_\theta,
\end{equation}
which is satisfied if $\rho_\theta$ solves the Zakai equation. This enables faster and more stable learning: because Eq. \ref{eq:PINO} enforces the governing equation, the resulting solutions are consistent with the known dynamics and training remains possible even when no labeled solutions exist, since for SPDEs the true filtering posterior is never observed. These models function by sampling collocation points $\{\mathbf{z}^{(k)}\}_{k=1}^K$ on the inferred domain, computing the derivatives of the network at those points through autodifferentiation, and minimizing the discrepancy between the two sides of Eq. \ref{eq:PINO} at each point.

\subsection{Model Architecture}
We use a unidirectional GRU \citep{cho2014propertiesneuralmachinetranslation, krishnan2015deepkalmanfilters} to encode the observation history into a causal context vector (Fig \ref{fig:1}. A, B) $\mathbf{v}_t \in \mathbb{R}^{d_v}$:
\begin{equation}
 \mathbf{h}_t = \operatorname{GRU}\big(\mathbf{h}_{t-1}, [\mathbf{x}_t \odot m_t,\, m_t]\big), \qquad \mathbf{v}_t = \mathbf{W}_\mathbf{v} \mathbf{h}_t + \mathbf{b}_\mathbf{v},
\end{equation} where $m_t \in \{0, 1\}$ indicates whether $\mathbf{x}_t$ is observed. By construction, $\mathbf{v}_t$ depends only on $X_{[0,t]}$. For each observation interval $[t, t + \Delta t]$, the causal context $\mathbf{v}_t$ is fed into a Deep Operator Network (DeepONet; App. \ref{app:DeepONet}) \citep{lu_learning_2021}. The DeepONet outputs the unnormalized log-density at any latent point $\mathbf{z}$ and any time within the interval \citep{bagmark_energy-based_2023, hu_score-based_2024}:
\begin{equation}
\log\rho_\theta(\mathbf{z}, t + s\Delta t \mid X_{[0,t]}) = \beta + b(\mathbf{v}_t, s)^\top\psi(\mathbf{z}), \qquad s \in [0, 1],
\end{equation} 
where $s$ is the normalized time since the observation at $t$ (i.e., $s = 0$ is the posterior after incorporating $\mathbf{x}_t$ and $s = 1$ is the prediction for $t + \Delta t$). The branch $b : \mathbb{R}^{d_v + 1} \to \mathbb{R}^p$ is an MLP with one hidden layer (tanh) while the trunk $\psi : \mathbb{R}^n \to \mathbb{R}^p$ is an MLP with three hidden layers (softplus). $\beta \in \mathbb{R}$ is a learned scalar bias. Because the trunk network can be queried at any $\mathbf{z}$ the model is mesh-free (Fig \ref{fig:1}. C).

We parameterize the drift $f_{\phi}(\mathbf{z})$ of equation \ref{eq:state_sde} as an MLP. We define the diffusion as a shared parameter $g_\phi$. For the readout, $h_\phi$ we use a Stiefel observation map \citep{Lee2026.03.07.710319} with learned diagonal observation noise,
\begin{equation}
     h_\phi(\mathbf{z}_t) = \mathbf{C}_\phi\mathbf{z}_t + \mathbf{d}_\phi + \epsilon \sim \mathcal{N}(0, \mathbf{R}_\phi), \qquad \mathbf{R}_\phi = \operatorname{diag}(r_1, \dots, r_m),
\end{equation} where $\mathbf{C}_\phi \in \mathbb{R}^{m \times n}$ has orthonormal columns ($\mathbf{C}_\phi^\top \mathbf{C}_\phi = \mathbf{I}_n$), i.e.\ lies on the Stiefel manifold \citep{Bendokat_2024}, $\mathbf{d}_\phi$ is a bias vector, and $\mathbf{R}_\phi$ is the observation noise covariance (Fig \ref{fig:1}D).

\paragraph{Initialization.} We initialize $\mathbf{C}_\phi$, $f_\phi$ and $\hat{\mathbf{z}}$ by subspace identification \citep{Van_Overschee_De_Moor_1996,qin_overview_2006} before training the operator (Appendix~\ref{app:init}) to start in a good initial estimate for learning.
\subsection{Fitting Procedure}
\label{sec:fitting}
We fit the model with an approach inspired by expectation-maximization \citep{Ghahramani_Roweis_1998}, alternating between learning $\log\rho_\theta$ with the dynamical and observation parameters $\phi = \{f, g, \mathbf{C}, \mathbf{d}, \mathbf{R}\}$ held fixed (E step) and refitting $\phi$ with $\rho_\theta$ frozen (M step). Full details are given in Appendix~\ref{app:fitting}.
 
\paragraph{E step.} We update $\theta$ by stochastic gradient descent on
\begin{equation}
    l_{E} = w_{\text{res}}\,l_{\text{res}} + l_{\text{jump}} + l_{\text{ic}},
\end{equation}
where each step draws a fresh batch of mesh-free collocation points from a defensive Gaussian mixture \citep{Hesterberg01051995}, half around the projection $\hat{ \mathbf{z}}_t = \mathbf{C}_\phi^\top(\mathbf{x}_t - \mathbf{d}_\phi)$ and half from a broad Gaussian at the origin. The residual loss $l_{\text{res}}$ is the scale-normalized squared residual of the Fokker--Planck equation (Eq.~\ref{eq:PINO}), written for $\log\rho_\theta$, between observation times. The jump loss $l_{\text{jump}}$ is a cross entropy that matches the posterior after each observation to the predicted density multiplied by the likelihood $p(\mathbf{x}_{t+1} \mid \mathbf{z}_{t+1})$, i.e., the Zakai measurement update. The initial condition loss $l_{\text{ic}}$ matches $p_\theta(\mathbf{z}_0 \mid \mathbf{x}_0)$ to the posterior under a $\mathcal{N}(0, \mathbf{I})$ prior. Densities are normalized locally over the collocation points by self-normalized importance sampling \citep{mcbook}(Appendix~\ref{app:estep}). After $N$ E-step updates, we freeze $\rho_\theta$ and compute the posterior means $\bar{\mathbf{z}_t} = \mathbb{E}_{p_\theta(\mathbf{z}_t \mid X_{[0,t]})}[\mathbf{z}_t]$ with the Metropolis-adjusted Langevin algorithm (MALA) \citep{Roberts_Tweedie_1996}
 
\paragraph{M step} After we have an estimate of $\bar{\mathbf{z}}$, we then fit $\phi$ in the following order:
\begin{enumerate}
    \item Fit the Stiefel readout $(\mathbf{C}_\phi, \mathbf{d}_\phi)$ by orthogonal Procrustes analysis \citep{Schönemann_1966}
    \item Regress the drift network $f_\phi$ on the increments of $\bar{\mathbf{z}_t}$ 
    \item Compute the scalar diffusion $g_\phi$ from the remaining increment residuals
    \item Estimate $\mathbf{R}_\phi$ from the observation directions orthogonal to $\operatorname{span}(\mathbf{C}_\phi)$
\end{enumerate}

\section{Results}
\subsection{Simulated Data} We first benchmark our model on three dynamical systems with known latents by collecting observations $\mathbf{x}_t \sim \mathcal{N}(\mathbf{C}^{*}\mathbf{z}_t + \mathbf{d}, \sigma_{obs}\mathbf{I})$ where $\mathbf{C}^*$ is a random linear projection with $\mathbf{C^*}_{ij} \sim \mathcal{N}(0,1)$. We evaluate our model on its latent and dynamical RMSE as well as its ability to reconstruct observations
\subsubsection{Latent Inference}

\begin{figure}[ht]
    \centering
\includegraphics[width=\linewidth,trim=0 6pt 0 0,clip]{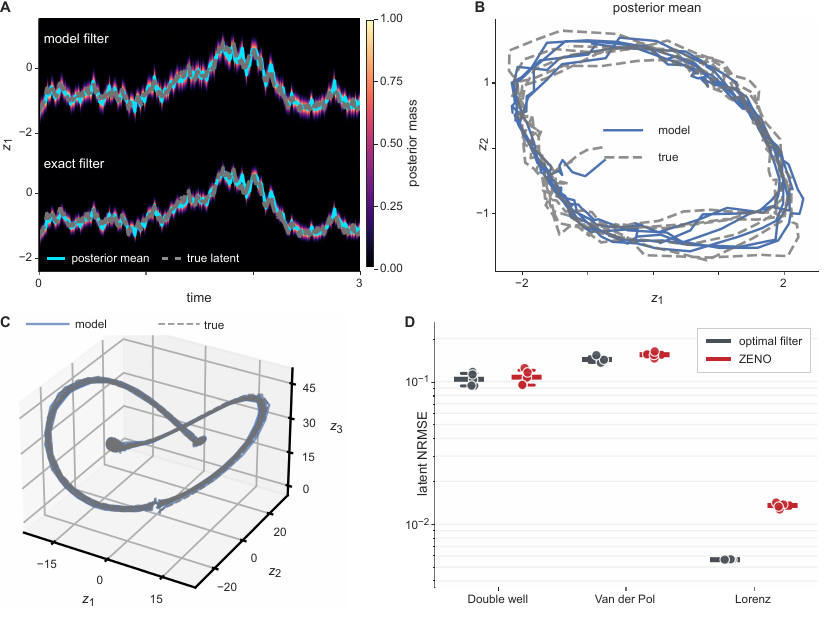}
    \caption{\textbf{Latent Inference of Simulated Dynamical Systems.} \textbf{A)} Exact filtering distribution for the double well system vs filtering posterior distribution estimated by ZENO. \textbf{B)} Latent Posterior Mean Estimate for the Van Der Pol System. \textbf{C)} Latent Trajectory Inference for the Lorenz Attractor. \textbf{D)} Comparison of ZENO's normalized root mean squared error with the exact optimal filter for each system.}
    \label{fig:latent_infernece}
\end{figure}

We first tested how well ZENO could infer the filtering posterior for a double well potential. This system is used in neuroscience as a simple model of decision making and is of interest because the posterior distribution can be bimodal \citep{pmlr-v97-duncker19a}. Because the state is low dimensional and we have access to the exact drift and diffusion, it is tractable to numerically solve the Zakai equation to access the Bayes optimal filter for comparison. In general, this would not be possible as numerical solvers scale exponentially with the dimension of the PDE they are solving. Fig \ref{fig:latent_infernece}A shows that we are able to represent the exact filtering distribution and its mean estimate while Fig \ref{fig:latent_infernece}B and C show that our model is able to scale to a higher dimensional system and accurately reconstruct the VDP oscillator and Lorenz system. We find that ZENO reaches the ceiling set by the optimal filter for the double well equation and Van Der Pol oscillator but gaps the optimal filter by about $0.005$ on the Lorenz system \cite{lorenz_deterministic_1963} (Fig. \ref{fig:latent_infernece}D). 

\subsubsection{Dynamics Parameter Learning}
\begin{figure}[!ht]
    \centering
     \includegraphics[width=\linewidth,trim=0 4pt 0 0,clip]{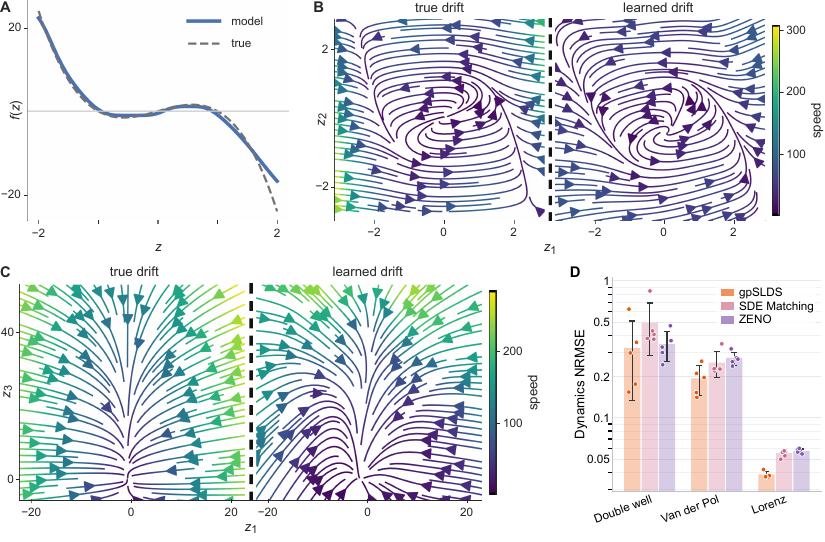}
    \caption{\textbf{ZENO Drift Estimation Performance.} \textbf{A)} True cubic drift of double well (dashed line) overlayed with model estimate. \textbf{B)} True Van Der Pol Vector field (Left) vs Model's infered Vector Field. \textbf{C)} Lorenz Attractor Dynamics plotted in the $xz$ plane. \textbf{D)} Model's NRMSE performance compared to baselines. Dots represent seeds and error bars are 1 SD.}
    \label{fig:dynamics}
\end{figure}

After we validated that latent inference by ZENO was accurate, we asked how well it learns system parameters. Fig \ref{fig:dynamics}A shows that ZENO learns the double well potential's cubic drift. On Van Der Pol, we show that ZENO reproduces the limit cycle but the velocity field is slower at the edges of the domain (Fig \ref{fig:dynamics}B). Finally, on Lorenz we find that ZENO reconstructs the dynamics but misestimates some directions and speeds of the flow field (Fig \ref{fig:dynamics}C). 

We compare ZENO to two established approaches: 1) the SING algorithm \citep{hu_sing_2025}, which fits a Gaussian process switching linear dynamical system (gpSLDS) \citep{hu2024modeling} to approximate the drift; and 2)SDE-matching, a simulation-free approach to variational inference of latent SDEs \citep{bartosh_sde_2025}. We found ZENO achieves similar levels of accuracy to these baselines: matching gpSLDS on doublewell and SDE matching on VDP (Fig \ref{fig:dynamics}D).

\subsection{Analysis of whole-brain calcium dynamics in {\textit{C. elegans}}}
\begin{figure}[!b]
    \centering
    \includegraphics[width=0.9\linewidth,trim=0 4pt 0 0,clip]{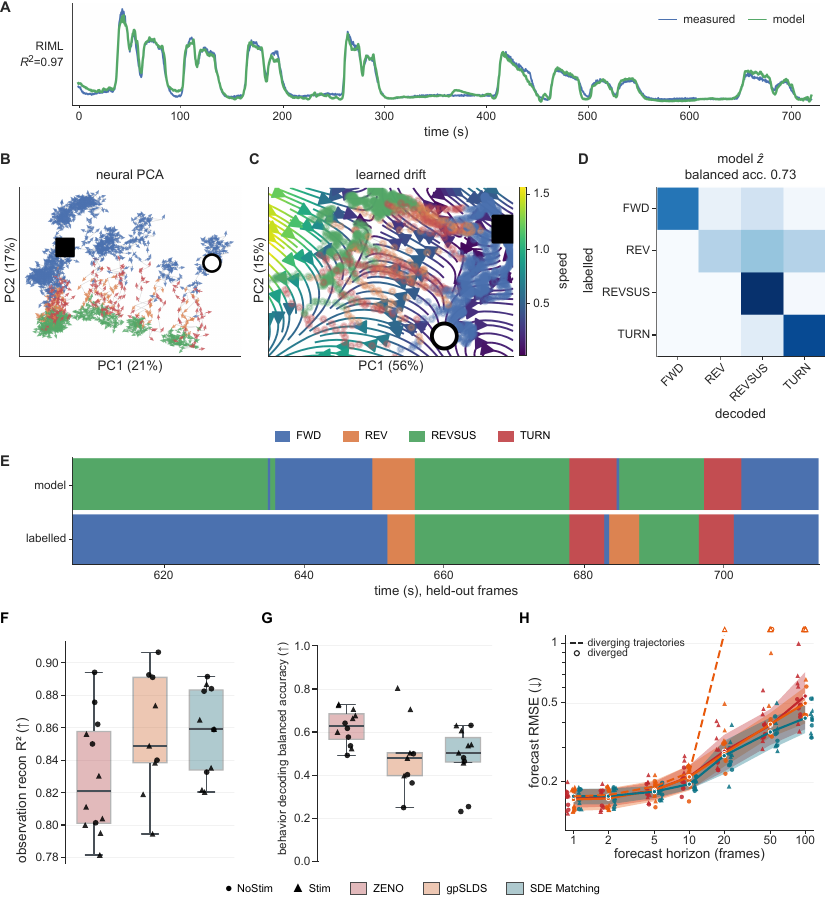}
    \caption{\textbf{Whole-brain calcium dynamics in {\textit{C. elegans}}} \textbf{A)} Example ZENO reconstruction of RIML neuron. \textbf{B-C)} ZENO recovers limit cycles. \textbf{D-E)} Inferred latents are consistent with known behavioral labels. Model performance compared to baselines on \textbf{F)} reconstruction $R^2$, \textbf{G)} behavior decoding accuracy, and \textbf{H)} forecasting RMSE. Points represent individual worms, separated into NoStim and Stim conditions. Error bars are 1SD.}
    \label{fig:kato}
\end{figure}

We used ZENO to analyze neuroscience data where the true latent state is unknown and where prior work \citep{linderman_hierarchical_2019,NEURIPS2020_aa1f5f73, yezerets_decomposed_2025} provides the opportunity to compare ZENO's latent state estimation with established methods. We tested ZENO on a dataset of whole-brain calcium dynamics in immobilized \textit{C. elegans} \citep{kato_global_2015}. Fig. \ref{fig:kato}A-E shows performance on an example worm (StimWorm 1). Specifically, Fig \ref{fig:kato}A shows that ZENO accurately 
reconstructs neural observations from the inferred latents. The latent states inferred by ZENO recapitulate reported limit cycles (Fig. \ref{fig:kato}B,C) and the inferred latents are behaviorally relevant and consistent with human-labels (Fig. \ref{fig:kato}D,E). We find that ZENO performs comparable to established state-of-the-art methods on reconstruction $R^2$ and forecasting RMSE\footnote{Forecasting was attempted with gpSLDS but led to inconsistent results likely due to hyperparameter tuning that we do not think accurate reflect the method.}. These results establish that ZENO performs comparably to established methods on real-world datasets of scientific value and can be used for scientific discovery of latent variables from neural measurments. 

\section{Related Work}
\textbf{Switching Linear Dynamical Systems.} Switching linear dynamical systems are hierarchical models that approximate a nonlinear system with mixtures of linear dynamical systems. They are often fit with Bayesian methods (e.g., expectation maximization) or with variational approximation. Their mixture of linear systems structure balances interpretability and tractability against expressivity: their discrete states and linear dynamics can be interpreted for scientific discovery \citep{linderman_bayesian_2017} but their approximate (variational) posteriors, while tractable fit to data, may diverge from the true posterior. Depending on the choice of variational family these models may struggle to match posteriors that are multimodal or asymmetric. ZENO forgoes a hierarchical framework in favor of directly approximating the true posterior distribution.

\textbf{Latent SDEs.} Latent SDEs address a similar problem using a variational autoencoder framework where the prior and posterior processes are governed by a neural stochastic differential equation and the corresponding drift and diffusion functions are modeled as neural networks \citep{pmlr-v108-li20i}. These models offer the flexibility and expressivity of neural networks, optimization with gradient descent, and parallelization offered by GPUs. However, balancing the data reconstruction and Kullback-Leibler (KL) loss that defines their variational objective can often be difficult, (e.g., posterior collapse, where the posterior equals the prior and becomes unusable for anything outside of reconstruction).
\textbf{Simulation-based Latent SDEs} established this approach first, using the stochastic adjoint sensitivity method. This required rollouts or simulations from the variational posterior at each gradient step to estimate a path-wise KL loss, which can be expensive for long timescales and numerically sensitive. \textbf{Simulation-free Latent SDEs} extend these ideas without simulation \citep{course_amortized_2023}. In SDE-matching, \cite{bartosh_sde_2025} use methods inspired by diffusion models and flow matching to learn latent dynamics through a KL divergence between the posterior and prior drift functions. In Neural Stochastic Flows, \cite{kiyohara2025neural} directly model the conditional law governing the solution to the latent SDE using normalizing flows.  Finally, in Neural Kolmogorov Equations, \cite{bizzi_towards_2025}, represents the SDE in its infinite dimensional form, but does not learn parameters of a latent SDE. ZENO is a simulation-free latent-SDE model that uses a physics based loss instead of a reconstruction/ELBO, enabling learning the parameters of a latent SDE.

\textbf{Physics-informed Neural Networks for the Fokker-Planck and Zakai Equations.} Other works have investigated using PINNs for solving the Fokker-Planck equation (FPE) \citep{xu_solving_2020, zhai2020deeplearningmethodsolving, chen_solving_2020, hu_score-based_2024} and Zakai equation \citep{bagmark_energy-based_2023, azimov_integrating_2026}. \cite{liu_solving_2021} solve the FPE as an inverse problem by fitting a network to the FPE solution and fitting the drift and diffusion in alternating steps. However, they use a grid based approach and only demonstrate it on one-dimensional systems. Furthermore, they assume that at least a noisy version of the PDF is accessible for training and do not handle unobserved latents. \cite{azimov_integrating_2026} learn the FPE solution and use it for Bayesian updates to solve for the filtering distribution, but assume known drift and diffusion functions. Both methods propagate the solution step-wise. To our knowledge, our approach is the first to simultaneously perform state inference of the full FPE solution operator while learning unknown dynamics functions using EM. 

\section{Discussion} Here, we introduced ZENO, a physics-informed neural operator that approximates the unnormalized filtering distribution by learning the solution operator of the Zakai equation. On synthetic systems, ZENO matched or came close to optimal numerical filters for latent inference and performed on par with state-of-the-art methods for parameter learning. On whole-brain \textit{C. elegans} recordings, it was comparable to baselines on forecasting and produced more behaviorally informative latents, at the cost of slightly lower reconstruction $R^2$, which we expect relaxing the Stiefel constraint on the observation model to address.

ZENO achieves similar performance to state-of-the-art methods but does not outperform them in all cases. All established methods we benchmarked against are estimating the smoothing distribution, which integrates observations both forward and backward in time, thereby providing a superior marginal distribution that can be used as a target. Future directions should investigate extending ZENO to inferring the smoothing distribution by learning the solution to the backwards stochastic PDE \citep{Pardoux01011982}. Additionally, it may be possible to improve the objective for fitting the drift and diffusion by casting it as an inverse problem but that is outside the scope of this work. Finally, our model uses a single set of parameters to represent the entire dynamical system even in cases where there are different dynamical regimes such as in our application to \textit{C. elegans}. Models such as decomposed linear dynamical systems \citep{yezerets_decomposed_2025} handle this through mixtures of dictionaries of linear dynamical systems. A similar idea could be combined with our model by using operator splitting methods \citep{serrano2026testtime}. In fact, one may be able to represent switches in regimes through jumps by generalizing the Forward Kolmogorov operator to Levy processes \citep{jia_njsde, leng2026deepzakaijstructuredfiltering}. Overall, ZENO represents a novel approach to the problem of nonlinear filtering and dynamical inference through the lens of partial differential equations and operator theory.
\subsection*{AI use statement}
In this work, we used generative AI tools for generation of synthetic data sets, to provide feedback on research methodology or experiments, and implement methods.
We have not used generative AI tools to propose or refine hypotheses, clean and reformat dataset, support qualitative and thematic data analysis, or interpret results and formulating mathematical claims, providing critical ingredients for proving mathematical claims, assisting in the writing of proofs, or assisting with translation are not applicable to this work. Additionally, we used generative AI tools for to modify scientific figures, suggest experimental parameters, create or edit software code, and brainstorming. We have reviewed all AI-assisted work. We
checked LLM-generated research ideas for potential plagiarism through a manual
literature survey. LLM-generated code was verified and tested for correctness by the authors. We take responsibility for the final content of this work,
including text, claims or artifacts produced with the aid of generative AI.

\subsection*{Reproducibility statement}
Code is opensourced at \url{https://github.com/aaprasad/zeno}. Model checkpoints, datasets and results assets can be provided upon request.
\subsubsection*{Author Contributions}
A.P. and B.D.D. devised the research question and model. A.P. implemented the model and ran experiments. A.P. and B.D.D wrote the paper. S.W.F. edited the paper and provided expertise on \textit{C. elegans} data. S.W.F. and B.D.D provided financial support for the project.
\ack{A.P is supported by the Henry E. Singleton Fellowship and the MIT Brain \& Cognitive Sciences Department. S.W.F is supported by the Howard Hughes Medical Institute and by the National Institute of Health (NIH R01 NS131457). B.D. is supported by a Alfred P. Sloan Foundation Research Fellowship.}

\bibliographystyle{plainnat}
\bibliography{references}

\newpage
\appendix
\section{Primer on Operator Learning}
\label{app:operator_primer}
\subsection{Formulation}
Most deep learning methods map between finite-dimensional vector spaces, $F:\mathbb{R}^n \to \mathbb{R}^m$. Many scientific problems instead involve maps between function spaces. For example, to solve a differential equation we are interested in solving
$$\mathcal{A}u=v$$ where $\mathcal{A}$ is a differential operator: a map that takes in a function and outputs its derivative. 
Formally, we want to approximate an operator

$$\mathcal{G}=\mathcal{A}^{-1}: \mathcal{V} \to \mathcal{U}, \qquad v \mapsto u = \mathcal{G}(v),$$

where $\mathcal{U}$ and $\mathcal{V}$ are (typically Banach) spaces of functions defined on domains $D \subset \mathbb{R}^{d}$ and $D' \subset \mathbb{R}^{d'}$. Operator learning fits a parametric approximation $\mathcal{G}_\theta \approx \mathcal{G}$ from input–output pairs $\{(u_i, v_i)\}_{i=1}^N$ by minimizing an error such as $\frac{1}{N}\sum_i \|\mathcal{G}_\theta(v_i) - u_i\|^2_{\mathcal{U}}$. An example of these input-output pairs could be the evaluations of a PDE on some domain paired with the PDE solution.

The difficulty is that functions are infinite-dimensional objects, while any computer sees them only through finite discretizations. While a high resolution discretization will lead to improved performance, generating high resolution data is expensive. A good operator-learning architecture defines a map that is meaningful independently of how inputs and outputs are sampled.

\subsection{Deep Operator Networks (DeepONet)}
\label{app:DeepONet}
DeepONet \citep{liu_solving_2021} builds on the universal approximation theorem for nonlinear operators of \cite{chen1995}. This theorem shows that any continuous operator can be approximated arbitrarily well by a sum of products between two networks: one that encodes the input function and one that encodes the output location.

Given an input function $v$ observed at $M$ fixed "sensor" locations $\{\mathbf{z}_m\}_1^M$, a DeepONet evaluates the output function at a query point $\mathbf{z} \in D'$ as
$$\mathcal{G}_\theta(v)(\mathbf{z}) = \sum_{k=1}^{p} b_k\big(v(\mathbf{z}_1),\dots,v(\mathbf{z}_M)\big)\, \psi_k(\mathbf{z}) + \beta .$$
The branch network $b:\mathbb{R}^M \to \mathbb{R}^p$ encodes the input function into $p$ coefficients while the trunk network $\psi:\mathbb{R}^{d'} \to \mathbb{R}^p$ produces $p$ basis functions evaluated at the query location. The solution operator is thus a linear combination of learned basis functions $\{\psi_k\}$ with input-dependent coefficients $\{b_k(v)\}$. This formulation enables the output to be mesh-free. The trunk can be queried at any $\mathbf{z}$, so outputs can be evaluated anywhere in $D'$. However, input is tied to a fixed discretization. The branch expects $v$ at the same $M$ sensors for every sample, so the input side is not discretization-invariant without further modification.

\section{Primer on Physics Informed Neural Networks (PINNs)}
\label{app:PINN_primer}
Classical numerical solvers for partial differential equations often fail to scale to higher dimensions due to exponential scaling with the dimensionality of the domain. Instead, modern machine learning based approaches amortize the solution to a PDE by directly fitting neural networks to the derivative-solution pairs. However, these data-driven models ignore known physics and tend to need large datasets which can be expensive to generate.
Physics-informed neural networks (PINNs) are a modern approach that represent the solution of a differential equation as a neural network and train it to satisfy both the observed data and the equation itself. This approach has been made practical by automatic differentiation in modern deep learning frameworks.

\subsection{Formulation}
Consider a general differential equation on a domain $\Omega \subset \mathbb{R}^d$, possibly with time $t \in [0,T]$:
\begin{align*}
    \mathcal{A}_\lambda u &= s(z,t), &&(\mathbf{z},t) \in \Omega \times (0,T], \\
    \mathcal{B}u &= b(\mathbf{z},t), &&(\mathbf{z},t) \in \partial\Omega \times (0,T], \\
    u(\mathbf{z},0) &= u_0(\mathbf{z}).
\end{align*}
Here, $\mathcal{A}_\lambda$ is a differential operator with parameters $\lambda$ (for example, $\mathcal{A}_\lambda = \partial_t - \mathcal{L}^*$ for the Fokker-Planck equation), and $\mathcal{B}$ is a boundary operator. A PINN approximates the solution by a network $u_\theta(\mathbf{z},t)$, typically a multilayer perceptron with smooth activations such as $\tanh$ or sinusoids.

Because $\mathcal{A}_\lambda[u_\theta]$ can be evaluated exactly at any point by automatic differentiation of the network with respect to its inputs, one defines
\begin{equation}
    \mathcal{A}_\lambda u_\theta(\mathbf{z},t) \approx s(\mathbf{z},t),
\end{equation}
which is true if $u_\theta$ solves the differential equation.

\subsection{Training}

Training minimizes a composite loss evaluated on sets of points sampled from the domain:

$$\mathcal{L}(\theta) = w_r \mathcal{L}_r + w_b \mathcal{L}_b + w_0 \mathcal{L}_0 + w_d \mathcal{L}_d,$$

with one term per source of information:

$$
\mathcal{L}_r = \frac{1}{N_r}\sum_{i=1}^{N_r} \big| 
\mathcal{A}_\lambda[u_\theta](\mathbf{z}_i^r, t_i^r) - s(\mathbf{z}_i^r,t_i^r) 
\big|^2, \qquad 
\mathcal{L}_b = \frac{1}{N_b}\sum_{i=1}^{N_b} \big| \mathcal{B}[u_\theta](\mathbf{z}_i^b, t_i^b) - b(\mathbf{z}_i^b,t_i^b) \big|^2,
$$
$$
\mathcal{L}_0 = \frac{1}{N_0}\sum_{i=1}^{N_0} \big| u_\theta(\mathbf{z}_i^0, 0) - u_0(\mathbf{z}_i^0) \big|^2, \qquad
\mathcal{L}_d = \frac{1}{N_d}\sum_{i=1}^{N_d} \big| u_\theta(\mathbf{z}_i^d, t_i^d) - u_i^{\text{obs}} \big|^2.
$$

$\{(\mathbf{z}_i^r,t_i^r)\}$ are called collocation points. The loss can be read as a soft penalty method: the differential equation acts as a regularizer that constrains the network wherever data are absent.
\section{ZENO Fitting Procedure}
\label{app:fitting}
This section gives the full E and M steps summarized in Section~\ref{sec:fitting}. We alternate between learning the unnormalized log-density $\ell_\theta = \log\rho_\theta$ with $\phi = \{f, g, \mathbf{C}, \mathbf{d}, \mathbf{R}\}$ fixed and refitting $\phi$ with $\rho_\theta$ frozen, in the spirit of expectation-maximization \citep{Ghahramani_Roweis_1998}. Throughout, $m_t \in \{0,1\}$ indicates whether $\mathbf{x}_t$ is observed, and the Gaussian log-likelihood is $\log p(\mathbf{x}_t \mid \mathbf{z}) = -\tfrac{1}{2}(\mathbf{x}_t - \mathbf{C}\mathbf{z} - \mathbf{d})^\top \mathbf{R}^{-1}(\mathbf{x}_t - \mathbf{C}\mathbf{z} - \mathbf{d})$ up to a constant. Hyperparameters are listed in Appendix~\ref{app:hparams_zeno}.
 
\subsection{E step}
\label{app:estep}
During the E step, we hold $\phi$ fixed and update the operator's parameters $\theta$ by gradient descent on $l_E$, recomputed at each step from a freshly sampled batch of collocation points.
 
\paragraph{Collocation points.} At every $(t, \text{trajectory})$ site we draw $K$ collocation points, half from a component around the data-driven latent estimate and half from a broad ``defensive'' component that covers regions the first might miss \citep{Hesterberg01051995}:
\begin{equation}
    \label{eq:app_q}
    q_t(\mathbf{z}) = \frac{1}{2}\mathcal{N}\big(\mathbf{z};\, m_t\hat{\mathbf{z}_t},\, \sigma_t^2 I\big) + \frac{1}{2}\mathcal{N}\big(\mathbf{z};\, 0,\, \sigma_b^2 I\big), \qquad \hat{\mathbf{z}_t} = \mathbf{C}_\phi^\top(\mathbf{z}_t - \mathbf{d}_\phi), \qquad \sigma_t = \begin{cases}\sigma_n & m_t = 1\\ \sigma_b & m_t = 0.\end{cases}
\end{equation}
At unobserved frames the first component is therefore also broad and centered at the origin.
 
\paragraph{Residual loss.} For the isotropic diffusion $g$, dividing the Fokker--Planck equation $\partial_t\rho = \mathcal{L}^*\rho$ by $\rho$ gives its log-space form $\partial_t\ell = \mathcal{A}_\phi[\ell]$ with
\begin{equation}
    \mathcal{A}_\phi[\ell](\mathbf{z}) = -\nabla\!\cdot f_\phi(\mathbf{z}) - f_\phi(\mathbf{z})\cdot\nabla\ell(\mathbf{z}) + \tfrac{1}{2}g_\phi^2\big(\lVert\nabla\ell(\mathbf{z})\rVert^2 + \Delta\ell(\mathbf{z})\big).
\end{equation}
Because $s$ is time normalized by the observation interval, $\partial_t = \Delta t^{-1}\partial_s$, and we minimize the scale-normalized residual
\begin{equation}
    \label{eq:app_res}
l_{\text{res}} = \frac{1}{|\mathcal{C}|}\sum_{(t,s,\textbf{z})\in\mathcal{C}} \frac{\big(\Delta t^{-1}\partial_s\ell_\theta - \mathcal{A}_\phi[\ell_\theta]\big)^2}{\operatorname{sg}\big[\mathcal{A}_\phi[\ell_\theta]\big]^2 + \operatorname{sg}\big[\Delta t^{-1}\partial_s\ell_\theta\big]^2 + 1},
\end{equation}
where $\operatorname{sg}(\cdot)$ is the stop-gradient operator and $\mathcal{C}$ contains, for each trajectory, a random subset of observation intervals $t$ redrawn at every step, $N_s$ evenly spaced times $s \in [0, 1]$ within each interval, and that interval's collocation points $\mathbf{z}$. The denominator normalizes the residual to prevent blow up at collocation points far from the mode of a sharp density \citep{doi:10.1137/20M1318043}, and stop-gradienting the denominator prevents the optimizer from reducing the loss by inflating that magnitude rather than by reducing the residual. The spatial derivatives of the trunk network, the time derivative of the branch network and the divergence of the drift are computed with forward-mode automatic differentiation.
 
\paragraph{Self-normalized densities.} The normalized log-posterior after the observation at $t$ is estimated over the collocation points by self-normalized importance sampling \citep{mcbook}:
\begin{equation}
    \label{eq:app_snis}
    \log p_\theta\big(\mathbf{z}_t^{(k)} \mid X_{[0,t]}\big) = \ell_\theta\big(\mathbf{z}_t^{(k)}, 0 \mid X_{[0,t]}\big) - \text{LSE}_k\big(w_t^{(k)}\big) + \log K,
\end{equation}
with importance weights $w_t^{(k)} = \ell_\theta\big(\mathbf{z}_t^{(k)}, 0 \mid X_{[0,t]}\big) - \log q_t\big(\mathbf{z}_t^{(k)}\big)$, where $\text{LSE}_k$ is the log-sum-exp over index $k$, so that $\frac{1}{K}\sum_k e^{w_t^{(k)}}$ estimates the normalizer. 
 
\paragraph{Jump loss.} The predicted density is represented by self-normalized weights on the next interval's collocation points,
\begin{equation}
    \label{eq:app_wpred}
    \log W_{\text{pred},t}^{(k)} = \text{logsoftmax}_k\Big(\text{sg}\Big[\ell_\theta\big(\mathbf{z}_{t+1}^{(k)}, 1 \mid X_{[0,t]}\big) - \log q_{t+1}\big(\mathbf{z}_{t+1}^{(k)}\big)\Big]\Big),
\end{equation}
where $\ell_\theta(\mathbf{z}_{t+1}^{(k)}, 1\mid X_{[0,t]})$ is interval $t$'s density propagated to $s=1$ using coefficients conditioned only on history through $t$, but evaluated at interval $t+1$'s collocation points. Multiplying by the likelihood gives the target of the Zakai measurement update,
\begin{equation}
    \label{eq:app_wtgt}
    \log W^{(k)}_{\text{tgt},t} = \text{logsoftmax}_k\Big(\log W_{\text{pred},t}^{(k)} + m_{t+1}\log p\big(\mathbf{x}_{t+1} \mid \mathbf{z}_{t+1}^{(k)}\big)\Big),
\end{equation}
so at an unobserved frame the target is the predicted density itself and the density simply propagates. We match the posterior at $t+1$ to this target with the cross entropy
\begin{equation}
    \label{eq:app_jump}
    l_{\text{jump}} = -\frac{1}{B(T-1)}\sum_{b}\sum_{t=0}^{T-2}\sum_k W^{(k)}_{\text{tgt},t} \log p_\theta\big(z_{t+1}^{(k)} \mid X_{[0,t+1]}\big),
\end{equation}
averaged over the $B$ trajectories.
 
\paragraph{Initial condition loss.} We match the first posterior to the prior $\mathcal{N}(0, I)$ multiplied by the first likelihood:
\begin{equation}
    \label{eq:app_ic}
    l_{\text{ic}} = -\frac{1}{B}\sum_b\sum_{k} W^{(k)}_{\text{ic}} \log p_\theta\big(\mathbf{z}_0^{(k)} \mid \mathbf{x}_0\big),
\end{equation}
where
\begin{equation}
    \label{eq:app_wic}
    \log W^{(k)}_{\text{ic}} = \text{logsoftmax}_k\Big(\log \mathcal{N}\big(\mathbf{z}_0^{(k)}; 0, I\big) + m_0\log p\big(\mathbf{x}_0 \mid \mathbf{z}_0^{(k)}\big) - \log q_0\big(\mathbf{z}_0^{(k)}\big)\Big).
\end{equation}
 
\paragraph{Total loss.} The E-step loss is
\begin{equation}
    \label{eq:app_le}
    l_{E} = w_{\text{res}}\,l_{\text{res}} + l_{\text{jump}} + l_{\text{ic}},
\end{equation}
where $w_{\text{res}} \geq 0$ balances the residual, which spreads the posterior, against the measurement updates, which sharpen it. The jump and initial-condition terms use every interval, while the residual uses the random subset of intervals described above. We minimize $l_E$ with Adam \citep{kingma_adam_2017} under an exponentially decaying learning rate; each update uses the gradient over all training trajectories, accumulated in chunks to bound memory, and resamples the collocation points and residual intervals, so the E step consists of many such updates rather than a single fit to convergence.
 
Algorithm~\ref{alg:estep} summarizes one E-step update.
 
\begin{algorithm}[t]
\caption{One gradient update to $\log\rho_\theta$}
\label{alg:estep}
\begin{algorithmic}[1]
\Require operator parameters $\theta$; fixed $\phi = \{f_\phi, g_\phi, \mathbf{C}_\phi, \mathbf{d}_\phi, \mathbf{R}_\phi\}$; training windows $X_{0:T-1}$ with observation mask $\mathbf{m}_{0:T-1}$ for $B$ trajectories; interval $\Delta t$
\Require collocation sizes $K$ (state), $N_s$ (time in interval), $N_t$ (residual intervals); proposal scales $\sigma_n, \sigma_b$; residual weight $w_{\text{res}}$
\State $\hat{\mathbf{z}_t} \gets \mathbf{C}_\phi^\top(\mathbf{x}_t - \mathbf{d}_\phi)$ for all $t$, $b$
\For{each site $(t, b)$}
    \State draw $\mathbf{z}_t^{(k)}$: half from $\mathcal{N}(m_t\hat{\mathbf{z}}_t, \sigma_t^2 \mathbf{I})$, half from $\mathcal{N}(0, \sigma_b^2\mathbf{I})$ \Comment{Eq.~\ref{eq:app_q}}
\EndFor
\State draw a random subset $\mathcal{T}$ of $N_t$ intervals; set $\mathcal{S} = \{0, \tfrac{1}{N_s - 1}, \dots, 1\}$
\State compute contexts $v_t$ with the causal GRU and $\ell_\theta(\mathbf{z}_t^{(k)}, s \mid X_{[0,t]})$ at $s \in \{0, 1\}$
\State $\log p_\theta(\mathbf{z}_t^{(k)} \mid X_{[0,t]}) \gets$ self-normalized estimate \Comment{Eq.~\ref{eq:app_snis}}
\For{$t = 0, \dots, T-2$}
    \State $\log W_{\text{pred},t}^{(k)} \gets \text{logsoftmax}_k\big(\operatorname{sg}[\ell_\theta(\mathbf{z}_{t+1}^{(k)}, 1 \mid X_{[0,t]}) - \log q_{t+1}(\mathbf{z}_{t+1}^{(k)})]\big)$ \Comment{Eq.~\ref{eq:app_wpred}}
    \State $\log W_{\text{tgt},t}^{(k)} \gets \text{logsoftmax}_k\big(\log W_{\text{pred},t}^{(k)} + m_{t+1}\log p(\mathbf{z}_{t+1} \mid \mathbf{z}_{t+1}^{(k)})\big)$ \Comment{Eq.~\ref{eq:app_wtgt}}
\EndFor
\State $l_{\text{jump}} \gets$ cross entropy of $\log p_\theta(\mathbf{z}_{t+1}^{(k)} \mid X_{[0,t+1]})$ under $W_{\text{tgt},t}$ \Comment{Eq.~\ref{eq:app_jump}}
\State $l_{\text{ic}} \gets$ cross entropy of $\log p_\theta(\mathbf{z}_0^{(k)} \mid \mathbf{x}_0)$ under $W_{\text{ic}}$ \Comment{Eqs.~\ref{eq:app_ic}--\ref{eq:app_wic}}
\For{$(t, s, \mathbf{z}) \in \mathcal{T} \times \mathcal{S} \times \{z_t^{(k)}\}$}
    \State forward-mode AD: $\partial_s\ell_\theta$, $\nabla\ell_\theta$, $\Delta\ell_\theta$, $f_\phi$, $\nabla\cdot f_\phi$; form $\mathcal{A}_\phi[\ell_\theta]$
\EndFor
\State $l_{\text{res}} \gets$ scale-normalized residual \Comment{Eq.~\ref{eq:app_res}}
\State $l_E \gets w_{\text{res}}\,l_{\text{res}} + l_{\text{jump}} + l_{\text{ic}}$ \Comment{Eq.~\ref{eq:app_le}}
\State $\theta \gets \text{Adam}(\theta, \nabla_\theta l_E)$, accumulating $\nabla_\theta l_E$ over chunks of trajectories
\end{algorithmic}
\end{algorithm}
\paragraph{Posterior means.} After $N$ gradient steps, we freeze $\theta$ and estimate $\bar{ \mathbf{z}_t} = \mathbb{E}_{p_\theta(\mathbf{z}_t\mid X_{[0,t]})}[\mathbf{z}_t]$ by sampling each filtering marginal $p_\theta(\mathbf{z}_t \mid X_{[0,t]}) \propto \exp \ell_\theta(\mathbf{z}, 0 \mid X_{[0,t]})$ with the Metropolis-adjusted Langevin algorithm (MALA) \citep{Roberts_Tweedie_1996}. For each (time, trajectory) site we run $K_c$ parallel chains, initialized at $\hat{ \mathbf{z}_t} = \mathbf{C}_\phi^\top(\mathbf{x}_t - \mathbf{d}_\phi)$ at observed frames and from $\mathcal{N}(0, \sigma_b^2 \mathbf{I})$ at unobserved frames, with the proposal
\begin{equation}
    \label{eq:app_mala}
    \mathbf{z}' = \mathbf{z} + \epsilon\,\nabla_\mathbf{z}\ell_\theta(\mathbf{z}, 0) + \sqrt{2\epsilon}\,\xi, \qquad \xi \sim \mathcal{N}(0, \mathbf{I}),
\end{equation}
followed by a Metropolis--Hastings correction. During the first third of the sweeps (burn-in) the step size $\epsilon$ of each site adapts toward an acceptance rate of $0.574$ \citep{roberts_rosenthal_98}, and $\bar{\mathbf{z}_t}$ is the average of the chain states over the remaining sweeps.
 
\subsection{M step}
\label{app:mstep}
Once, we have a posterior mean estimate, we refit $\phi$ through an M step.
 
\paragraph{Observation map.} We first refit the Stiefel readout by orthogonal Procrustes analysis \citep{Schönemann_1966}:
\begin{equation}
    \label{eq:app_procrustes}
    \mathbf{C}_\phi = \mathbf{U}\mathbf{V}^\top, \qquad \mathbf{d}_\phi = \bar{\mathbf{x}} - \mathbf{C}_\phi \bar{\mathbf{z}},
\end{equation}
where $\mathbf{U}\mathbf{V}^\top$ is the singular value decomposition of the empirical cross-covariance $M = \sum_t (\mathbf{x}_t - \bar{\mathbf{x}})(\bar{ \mathbf{z}}_t - \bar{\mathbf{z}})^\top$ and $\bar x$, $\bar{\mathbf{z}}$ are the means of the observations and of the posterior means. We record how far the readout moved, $\delta_{\mathbf{C}_\phi} = 1 - \frac{1}{n}\sum_i \cos\vartheta_i$, where $\vartheta_i$ are the principal angles between the new and previous column spaces of $\mathbf{C}_\phi$.
 
\paragraph{Drift.} Only if $\delta_{\mathbf{C}_\phi}$ is below a tolerance, so that the latent coordinates have stabilized, do we refit the drift, by minimizing with Adam, warm-started from the current network,
\begin{equation}
    \label{eq:app_drift}
 l_{\text{drift}} = \frac{1}{|\mathcal{V}|}\sum_{t\in\mathcal{V}} \Big\lVert f_\phi(\tilde{ \mathbf{z}_t}) - \frac{\bar{ \mathbf{z}_{t+1}} - \bar{ \mathbf{z}_t}}{\Delta t}\Big\rVert^2 + \lambda\lVert\phi_f\rVert^2 + \alpha\, l_{\text{curv}},
  \qquad \tilde{\mathbf{z}_t} = \bar{\mathbf{z}} + \tfrac{\Delta t}{2} f_\phi^{\text{old}}(\bar{\mathbf{z}_t}),
\end{equation}
where $\mathcal{V}$ is the set of times at which both $\mathbf{x}_t$ and $\mathbf{x}_{t+1}$ are observed and $f_\phi^{\text{old}}$ is the drift before the refit. Evaluating the drift at the second-order Runge--Kutta midpoint $\tilde{\mathbf{z}_t}$ removes the $O(\Delta t)$ bias of regressing $f_\phi(\bar{ \mathbf{z}_t})$ on the forward increment. The weight decay $\lambda\lVert\phi_f\rVert^2$ acts on all weights and biases, and the curvature penalty is a finite-difference estimate of the squared second directional derivative along a random unit direction $\mathbf{u}_t$, redrawn at each iteration,
\begin{equation}
    \label{eq:app_curv}
    l_{\text{curv}} = \frac{1}{\delta^4|\mathcal{V}|}\sum_{t\in\mathcal{V}}\big\lVert f_\phi(\tilde{\mathbf{z}_t} + \delta \mathbf{u}_t) - 2f_\phi(\tilde{\mathbf{z}_t}) + f_\phi(\tilde{\mathbf{z}_t} - \delta \mathbf{u}_t)\big\rVert^2,
\end{equation}
with $\delta$ equal to $5\%$ of the average standard deviation of the latent coordinates.
 
\paragraph{Diffusion.} We refit the scalar diffusion from the trapezoidal increment residual under the current drift:
\begin{equation}
    \label{eq:app_g}
g_\phi = \max\Big\{\Big(\frac{\Delta t}{n|\mathcal{V}|}\sum_{t\in\mathcal{V}}\lVert r_t\rVert^2\Big)^{1/2},\, g_{\min}\Big\}, \qquad r_t = \frac{\bar{ \mathbf{z}_{t+1}} - \bar{ \mathbf{z}_t}}{\Delta t} - \tfrac{1}{2}\big(f_\phi(\bar{\mathbf{z}}) + f_\phi(\bar{ \mathbf{z}_{t+1}})\big).
\end{equation}
 
\paragraph{Observation noise.} Finally, we refit the diagonal observation noise $\mathbf{R}_\phi = \operatorname{diag}(r_1, \dots, r_m)$ from the observation directions orthogonal to $\operatorname{span}(\mathbf{C}_\phi)$, which no latent state can explain:
\begin{equation}
    \label{eq:app_R}
r_i = \frac{1}{|\mathcal{O}|}\sum_{t \in \mathcal{O}} \frac{\big[(I - \mathbf{C}_\phi \mathbf{C}_\phi^\top)(\mathbf{x}_t - \mathbf{d}_\phi)\big]_i^2}{1 - \lVert {\mathbf{C}_{\phi}}_{i,:}\rVert^2},
\end{equation}
where $\mathcal{O}$ is the set of observed frames and $1 - \lVert {\mathbf{C}_{\phi}}_{i,:}\rVert^2 = (\mathbf{I} - \mathbf{C}_\phi \mathbf{C}_\phi^\top)_{ii}$.
 
Algorithm~\ref{alg:mstep} summarizes the M step, which is run once every $N$ E-step updates and once, in bootstrap mode, at initialization (Appendix~\ref{app:init}).
 
\begin{algorithm}[t]
\caption{M step}
\label{alg:mstep}
\begin{algorithmic}[1]
\Require frozen operator $\theta$; current $\phi = \{f_\phi, g_\phi, \mathbf{C}_\phi, \mathbf{d}_\phi, \mathbf{R}_\phi\}$; training windows $X_{0:T-1}$, mask $m_{0:T-1}$; interval $\Delta t$
\Require broad scale $\sigma_b$; stability tolerance $\tau_C$; drift iterations $M$, weight decay $\lambda$, curvature weight $\alpha$; floor $g_{\min}$; flag \textsc{bootstrap}
\Require MALA estimate of $\bar{\mathbf{z}}$
\State $\hat{\mathbf{z}_t} \gets \mathbf{C}_\phi^\top\mathbf{x}_t - \mathbf{d}_\phi)$ for all $t$
\If{\textsc{bootstrap}}
    \State $\bar{\mathbf{z}_t} \gets \hat{ \mathbf{z}_t}$ \Comment{no operator yet}
\EndIf
\State $\mathbf{C}_{\text{old}} \gets \mathbf{C}_\phi$; $(\mathbf{C}_\phi, \mathbf{d}_\phi) \gets$ orthogonal Procrustes fit of $\mathbf{x}$ on $\bar{\mathbf{z}}$ \Comment{Eq.~\ref{eq:app_procrustes}}
\State $\delta_\mathbf{C} \gets 1 - \frac{1}{n}\sum_i \cos\vartheta_i(\mathbf{C}_\phi, \mathbf{C}_{\text{old}})$
\State $\mathcal{V} \gets \{t : m_t = m_{t+1} = 1\}$
\If{$\delta_\mathbf{C} < \tau_\mathbf{C}$} \Comment{refit dynamics only once the sensor is stable}
    \State $\tilde {\mathbf{z}}_t \gets \bar{\mathbf{z}_t} + \tfrac{\Delta t}{2} f_\phi(\bar{mathbf{z}_t})$ for $t \in \mathcal{V}$ \Comment{RK2 midpoint, current drift}
    \For{$j = 1, \dots, M$}
        \State draw unit directions $\mathbf{u}_t$; $f_\phi \gets \text{Adam}(f_\phi, \nabla l_{\text{drift}})$ \Comment{Eqs.~\ref{eq:app_drift}--\ref{eq:app_curv}}
    \EndFor
\EndIf
\State $g_\phi \gets$ trapezoidal residual estimate, floored at $g_{\min}$ \Comment{Eq.~\ref{eq:app_g}}
\If{not \textsc{bootstrap}}
    \State $r_i \gets$ orthogonal-complement estimate for each dimension $i$ \Comment{Eq.~\ref{eq:app_R}}
\EndIf
\State \Return $\phi = \{f_\phi, g_\phi, \mathbf{C}_\phi, \mathbf{d}_\phi, \mathbf{R}_\phi\}$
\end{algorithmic}
\end{algorithm}
 
\subsection{Initialization}
\label{app:init}
To initialize the model, we take inspiration from subspace identification \citep{Van_Overschee_De_Moor_1996,qin_overview_2006}. We form the mean-centered future block Hankel matrix of the observations, take its top-$n$ left singular vectors, and set $\mathbf{C}_\phi$ to the orthonormalized first block row, with $\mathbf{d}_\phi = \bar x$. A linear dynamics matrix $\mathbf{A} = (\mathbf{A}_{\text{disc}} - \mathbf{I})/\Delta t$ is then obtained by least squares, $\hat{ \mathbf{z}_{t+1}} \approx \mathbf{A}_{\text{disc}}\hat{\mathbf{z}_t}$, on the projection $\hat{\mathbf{z}}_t = \mathbf{C}_\phi^\top(\mathbf{z}_t - \mathbf{d}_\phi)$. We warm start the drift network by regressing $f_\phi(\mathbf{z})$ onto $\mathbf{A}\mathbf{z}$ at points drawn uniformly from a box around the origin, initialize $g_\phi$ and $\mathbf{R}_\phi$ to fixed values, and initialize the operator randomly. Before any operator training we take one bootstrap M step that uses $\hat{\mathbf{z}}_t$ in place of the MALA posterior means; it refits $\mathbf{C}_\phi$, $\mathbf{d}_\phi$, $f_\phi$ and $g_\phi$ as above and leaves $\mathbf{R}_\phi$ at its initial value.

\section{Experimental Details}
\label{app:experimental_details}
All experiments are trained on one GPU; Models are initialized randomly and a new simulated dataset is generated per seed. However methods using the same seed use the same dataset as input. ZENO and all baselines are implemented in Python ($\geq$ 3.10) on JAX ($\geq$ 0.6) (\cite{bradbury_jax_2018}): ZENO uses Equinox ($\geq$ 0.13) for its networks and Optax ($\geq$ 0.2) for optimization \citep{deepmind2020jax}, SDE Matching uses the same stack with Diffrax (0.7) (\cite{kidger2021on}) for its forecast rollouts, gpSLDS uses the authors' SING implementation (Hu et al., 2025), and experiments are configured with Hydra \citep{Yadan2019Hydra}.

\subsection{Datasets}
\label{app:datasets}
\paragraph{Simulated systems.} We benchmark on latent SDEs $dz = f(z)\,dt + G\,dW$ with three drifts. For the double well and the Van der Pol oscillator we follow the systems of \citet{pmlr-v97-duncker19a}; for the Lorenz system we follow the setup of \citet{pmlr-v108-li20i, bartosh_sde_2025, hu_sing_2025}, replacing their multiplicative noise with an additive diffusion.
\begin{align*}
    \text{Double well:}\quad & f(z) = a\,z\,(1 - z^2), \quad a = 4, \\
    \text{Van der Pol:}\quad & f_1(\mathbf{z}) = \tau\mu\big(z_1 - \tfrac{1}{3}z_1^3 - z_2\big), \quad f_2(\mathbf{z}) = \tau z_1/\mu, \\
    & \tau = 15,\ \mu = 2, \\
    \text{Lorenz:}\quad & f_1(\mathbf{z}) = \sigma(z_2 - z_1), \quad f_2(\mathbf{z}) = z_1(\rho - z_3) - z_2, \quad f_3(\mathbf{z}) = z_1 z_2 - \beta z_3, \\
    & (\sigma,\rho,\beta) = (10, 28, 8/3).
\end{align*}
Each trajectory is simulated with an Euler--Maruyama scheme whose step is at most $1$\,ms and is refined so that every observation time is hit exactly. Observations are $\mathbf{x}_t \sim \mathcal{N}(\mathbf{C}\mathbf{z}_t + d, \epsilon^2\mathbf{I})$ on a regular grid. For the double well and Van der Pol, $C$ and $d$ are a random Gaussian sensor with elements drawn IID from the standard normal distribution and shared by all splits. the noise is added in physical units. For Lorenz, observations are $x_t \sim \mathcal{N}(\mathbf{z}_t, \epsilon^2 I)$, the clean signal is standardized with train-only constants, and the noise is added after this normalization, following the \texttt{diffrax} \cite{kidger2021on} reference. Training, validation and test trajectories are independent draws that share the sensor. Before training, observations are standardized using training data only through a per-channel centering and one global scale set by the leading singular values of the training observations. Table~\ref{tab:sim_datasets} lists the parameters of each dataset.

\begin{table}[htbp]
\centering
\caption{Simulated datasets. $d$ is the latent dimension and $D$ the number of observed channels. $G$ multiplies $dW$. The observation noise SD $\epsilon$ is in physical units for the double well and Van der Pol and in units of the standardized clean signal for Lorenz. Trajectory counts are train / validation / test.}
\label{tab:sim_datasets}
\small
\fitwidth{\begin{tabular}{lccc}
\toprule
 & Double well & Van der Pol & Lorenz \\
\midrule
Latent dimension $d$ & 1 & 2 & 3 \\
Drift parameters & $a = 4$ & $\tau = 15,\ \mu = 2$ & $(10, 28, 8/3)$ \\
Diffusion $G$ & $1$ & $I_2$ & $0.3\, I_3$ \\
Initial state $z_0$ & $\mathcal{N}(0, 1)$ & $\mathrm{Unif}([-3, 3]^2)$ & $\mathcal{N}(0, I_3)$ \\
Duration $T$ (s) & 3 & 3 & 2 \\
Observations per trajectory & 301 & 301 & 100 \\
Observation interval $\Delta t$ (s) & 0.01 & 0.01 & $2/99 \approx 0.0202$ \\
Simulation step (s) & $10^{-3}$ & $10^{-3}$ & $\le 10^{-3}$ \\
Sensor & Gaussian $C, d$ & Gaussian $C, d$ & identity \\
Observed channels $D$ & 15 & 20 & 3 \\
Observation noise SD $\epsilon$ & 0.5 & 1.5 & 0.01 \\
Trajectories (train / val / test) & 20 / 5 / 10 & 20 / 5 / 10 & 1024 / 128 / 128 \\
\bottomrule
\end{tabular}}
\end{table}

\paragraph{Whole-brain calcium imaging in \textit{C. elegans}.} The dataset of \citet{kato_global_2015} contains 12 immobilized worms. Five worms (WT NoStim) were recorded spontaneously for about 18 minutes (3021--3311 frames at 2.8--3.1\,Hz, 109--135 identified neurons), and seven worms (WT Stim) were recorded for about 12 minutes (2017--2201 frames, 107--151 neurons) while oxygen stimuli were delivered; we do not model the stimulus, so the Stim recordings are treated as an unconditional benchmark. We use the corrected fluorescence traces at the native frame rate with no smoothing, clipping or derivative preprocessing. Following the whole-trace protocol of \citet{yezerets_decomposed_2025}, we fit a separate model to each worm and fit and score every frame of the recording; reconstruction and forecasting scores on this dataset are therefore in-sample. Recordings are cut into windows of 200 frames (training stride 100, non-overlapping scoring windows); each neuron is centered by its training mean and all neurons are divided by one global signal scale, the geometric mean of the top-10 singular values of the training frames. Every method uses a 10-dimensional latent state and an observation-noise initialization of $0.1$ in standardized units, which is then learned. Human-annotated behavioral states accompany each recording (7 labeled states for NoStim, 4 for Stim; frames labeled \texttt{NOSTATE} are excluded); they are never used to fit or select a dynamical model.

\subsection{Training}
\label{app:training}
Every method receives the same standardized data, observation grid, dimensions and validation cadence. Gradient-based methods (ZENO, SDE Matching) train for 10{,}000 steps on the double well, Van der Pol and Kato datasets and 3{,}000 steps on Lorenz, with validation every 100 steps; gpSLDS is fit by variational EM for 50 iterations. Each method selects its checkpoint on the validation split with a criterion that does not use ground truth (Section~\ref{app:hparams_zeno}--\ref{app:hparams_sdem}), and the test split is scored once, after selection, with 128 posterior and forecast samples. Baselines run in float64; ZENO in float32. For the simulated systems we ran six seeds (0--5) per dataset and method; the seed changes both the data draw and the initialization. For Kato we fit one model per worm (seed 0) and treat animals as the replicates.

\subsubsection{ZENO}
\label{app:hparams_zeno}
The E-step minimizes $l_E$ with Adam \citep{kingma_adam_2017} with $\lambda=2\times10^{-3}$ with an exponential decay of $\beta = 0.9998$ per step, using full-batch gradients accumulated over chunks of 16 trajectories. The Fokker--Planck residual is evaluated at 4 time-in-interval collocation points $s \in [0, 1]$ on 24 randomly chosen observation intervals per step, and at $N_z$ mesh-free state collocation points per (interval, trajectory) drawn from the defensive mixture with $\sigma_n = 0.3$ and $\sigma_b = 1.6$. Every 2{,}000 E-steps we run one M-step: the filter mean is read out with 64 MALA chains of 30 sweeps per site (the first third is burn-in; the step size self-tunes to an acceptance rate of $0.574$ \citep{roberts_rosenthal_98}), the Stiefel sensor is refit by orthogonal Procrustes, the drift network is refit for 400 Adam iterations at learning rate $2\times10^{-3}$ against the increment target evaluated at the RK2 predicted midpoint (\texttt{det\_mid}), the scalar diffusion is refit from the trapezoidal increment residual, and the diagonal observation noise $R$ is refit from the residual directions orthogonal to $\mathrm{span}(C)$. The drift is refit only when the sensor has moved little since the previous M-step. The sensor and a linear dynamics matrix are initialized by temporal subspace identification; the drift network is warm-started to those linear dynamics and one bootstrap M-step from the projection $\hat{z} = C^\top(x - d)$ precedes any operator training. The operator itself is randomly initialized with the default uniform fan-in scheme. Checkpoints are selected by the forecast RMSE on the validation windows (16 forecast samples), the same quantity that is reported on the test split. Table~\ref{tab:hparams_zeno} lists the different ZENO hyperparameters used for each dataset.

\begin{table}[htbp]
\centering
\caption{ZENO hyperparameters. Shared values apply to every dataset; the last four rows are the only per-dataset settings.}
\label{tab:hparams_zeno}
\small
\fitwidth{\begin{tabular}{llc}
\toprule
Component & Hyperparameter & Value \\
\midrule
Context encoder & causal GRU hidden / context dimension & 64 / 64 \\
DeepONet & basis size $p$ & 64 \\
 & branch MLP hidden width & 128 \\
 & trunk MLP width $\times$ depth, activation & $64 \times 3$, softplus \\
Drift $f_\phi$ & MLP width $\times$ depth, activation & $64 \times 3$, tanh \\
Diffusion & scalar $g$: initial value / lower bound & 1.0 / 0.05 \\
Observation model & Stiefel $C$, bias $d$, diagonal $R$ (learned) & yes \\
E-step & optimizer, learning rate, decay per step & Adam, $2\times10^{-3}$, 0.9998 \\
 & $s$-collocation points, intervals per step & 4, 24 \\
 & collocation proposal $(\sigma_n, \sigma_b)$ & (0.3, 1.6) \\
 & residual weight $w_{\text{res}}$ & 0.4 \\
 & trajectories per gradient chunk & 16 \\
M-step & E-steps between M-steps & 2{,}000 \\
 & drift iterations, learning rate & 400, $2\times10^{-3}$ \\
 & drift weight decay $\lambda$ & $3\times10^{-4}$ \\
 & drift target & RK2 midpoint (\texttt{det\_mid}) \\
 & MALA chains $\times$ sweeps & $64 \times 30$ \\
 & sensor-stability tolerance for refitting $f$ & 0.05 \\
Initialization & sensor / dynamics & subspace ID + bootstrap M-step \\
Selection & criterion (validation), forecast samples & prefix forecast RMSE, 16 \\
\midrule
 & & DW / VDP / Lorenz / Kato \\
\midrule
Per dataset & latent $d$, observed $D$ & 1, 15 / 2, 20 / 3, 3 / 10, $N$ \\
 & state collocation points $N_z$ & 128 / 256 / 384 / 128 \\
 & drift curvature penalty $\alpha$ & $10^{-3}$ / $10^{-3}$ / $10^{-3}$ / 0 \\
 & training steps & 10k / 10k / 3k / 10k \\
\bottomrule
\end{tabular}}
\end{table}

\subsubsection{gpSLDS}
\label{app:hparams_gpslds}
We use the authors' SING implementation \citep{hu_sing_2025} of the gpSLDS \citep{hu2024modeling} with the smoothly switching linear (SSL) kernel. Table~\ref{tab:hparams_gpslds} lists the settings we used; the per-dataset entries were chosen on validation data from single-seed tuning runs. Held-out inference runs SING's E-step with the learned drift frozen, starting each trial from the closed-form posterior under the Brownian prior followed by drift-free E-steps (\texttt{warm\_iters}), which is what keeps inference stable at small $\sigma$; the variational inducing distribution is initialized at the prior $\mathcal{N}(0, K_{zz})$. On Kato, 3 of the 12 fits used a second configuration (a fixed prior noise $\sigma = 1$, 3 regimes, learned emission and SING's \texttt{neural} natural-gradient schedule) and the configuration reported for each animal is the one with the lower in-sample forecast RMSE. gpSLDS is therefore a per-animal best-of over hyperparameters on Kato, while ZENO and SDE Matching use one fixed configuration for all worms.

\begin{table}[htbp]
\centering
\caption{gpSLDS hyperparameters. The upper block is shared; the lower block is per dataset. Kato lists the primary configuration (9 of 12 animals; the alternative used for the other three is given in the text).}
\label{tab:hparams_gpslds}
\small
\fitwidth{\begin{tabular}{lc}
\toprule
Shared hyperparameter & Value \\
\midrule
Kernel & smoothly switching linear (SSL) \\
Variational EM iterations & 50 \\
Natural-gradient E-steps per iteration & 15 \\
Adam M-steps per iteration (kernel hyperparameters) & 50 (Kato: 10) \\
Drift-free warm E-steps per trial & 15 \\
Held-out inference iterations (parameters frozen) & 15 \\
Initial slope-variance prior & 1.0 \\
Regime fixed-point initialization & $\mathcal{N}(0, I)$ \\
Emission $C, d, R$ & closed-form updates (Kato: frozen at PCA init.) \\
Precision & float64 \\
\bottomrule
\end{tabular}}

\medskip
\fitwidth{\begin{tabular}{lcccc}
\toprule
Per-dataset hyperparameter & Double well & Van der Pol & Lorenz & Kato \\
\midrule
Linear regimes & 3 & 3 & 3 & 4 \\
Prior noise scale $\sigma$ (fixed by SING) & 1.0 & 1.0 & 0.3 & 0.3 \\
M-step learning rate & $10^{-2}$ & $3\times10^{-3}$ & $10^{-2}$ & $10^{-2}$ \\
SSL kernel expectation & 1-sample MC & 5-pt Gauss--Hermite & 3-pt Gauss--Hermite & $2d$ sigma points \\
Inducing points & 16 (grid) & $8 \times 8$ (grid) & $4^3$ (grid) & 64 ($k$-means) \\
Inducing-grid padding (fraction of PCA range) & 0.25 & 0.25 & 0.25 & -- \\
Natural-gradient schedule $\rho$ & synthetic & synthetic & constant 0.1 & constant 0.1 \\
Inference sub-steps per observation interval & 1 & 1 & 4 & 1 \\
Partition temperature $\tau$ & 0.5 & 0.5 & 0.5 & 5 \\
Trials per vEM iteration & 20 & 20 & 16 & all windows \\
\bottomrule
\end{tabular}}
\end{table}

\subsubsection{SDE Matching}
\label{app:hparams_sdem}
We implement the diagonal-Gaussian marginal specialization of SDE Matching \citep{bartosh_sde_2025} in JAX. The prior is a latent SDE with a neural drift and a constant diagonal diffusion; the posterior marginals $q(z_t \mid x) = \mathcal{N}(m(t), \mathrm{diag}\,s(t)^2)$ are produced by a network that reads the time and the state of a backward GRU run over the observations (so the encoder is a smoother), and the posterior drift follows in closed form from $dm/dt$, $ds/dt$ and the diffusion (time derivatives by forward-mode differentiation). No SDE is solved during training. The objective is the initial KL plus the path energy, estimated at 4 uniformly sampled times per trajectory and multiplied by the physical duration, plus the observation negative log-likelihood at sampled times multiplied by the number of observations; there is no KL annealing or time normalization. Forecasts integrate the learned prior from the posterior at the end of the prefix with an Euler--Maruyama step of $0.01$. Table~\ref{tab:hparams_sdem} lists the hyperparameters, which are identical on every dataset except for the step budget.

\begin{table}[htbp]
\centering
\caption{SDE Matching hyperparameters (identical for all datasets unless noted).}
\label{tab:hparams_sdem}
\small
\fitwidth{\begin{tabular}{ll}
\toprule
Hyperparameter & Value \\
\midrule
Prior drift network & MLP $d \to 64 \to 64 \to d$, tanh, last layer zero-initialized \\
Prior diffusion & constant diagonal, softplus-parameterized, initialized at 0.3 \\
Encoder & backward GRU, hidden size 64 \\
Marginal network $(m, \log s)$ & MLP $(2 \cdot 64 + 1) \to 64 \to 64 \to 2d$, tanh \\
Initial prior $p(z_0)$ & learned diagonal Gaussian \\
Emission $C, d$ & affine, initialized by PCA of the training observations \\
Observation noise & diagonal, learned; init.\ at the dataset noise SD (Kato: 0.1) \\
Matching times per trajectory & 4 \\
Optimizer, learning rate & Adam, $10^{-3}$ \\
Gradient-norm clipping & 10 \\
Batch size (trajectories) & 16 \\
Training steps & 10{,}000 (Lorenz: 3{,}000) \\
Validation cadence / selection & every 100 steps / validation objective (neg.\ ELBO) \\
Forecast solver step & 0.01 \\
Precision & float64 \\
\bottomrule
\end{tabular}}
\end{table}

\subsection{Evaluation Metrics}
\label{app:metrics}
\subsubsection{Simulated Data}
Because the generating latents are known, we score the simulated systems directly, on independent test trajectories. The latent state is identifiable only up to an affine transformation, so we first fit an unrestricted affine map $z \mapsto zA + b$ by least squares from the inferred posterior means to the true latents \emph{on the validation split}, after checkpoint selection, and freeze it before scoring the test split. This removes coordinate ambiguity without fitting away test error. No latent alignment enters the observation-space scores.
\paragraph{Reconstruction $R^2$ and RMSE.} With $\hat{\mathbf{x}}_t = \mathbf{C}_\phi\bar{\mathbf{z}}_t + \mathbf{d}_\phi$ the reconstruction from the posterior mean and $\mathbf{x}^\star_t = \mathbf{C}^*\mathbf{z}_t + \mathbf{d}^*$ the noiseless signal (both in standardized units),
\begin{equation*}
    \text{RMSE}_{\text{rec}} = \sqrt{\tfrac{1}{TBD}\textstyle\sum_{t,b,i} (\hat{\mathbf{x}}_{t,b,i} - \mathbf{x}^\star_{t,b,i})^2}, \qquad
    R^2_{\text{rec}} = 1 - \frac{\sum_{t,b,i} (\hat{\mathbf{x}}_{t,b,i} - \mathbf{x}^\star_{t,b,i})^2}{\sum_{t,b,i} (x^\star_{t,b,i} - \bar{\mathbf{x}}^\star_i)^2},
\end{equation*}
where $b$ indexes test trajectories, $i$ channels and $\bar{\mathbf{x}}^\star_i$ is the channel mean over the test split.
\paragraph{Latent NRMSE.} With $\tilde{\mathbf{z}}_t = \bar{\mathbf{z}}_t \mathbf{A} + \mathbf{b}$ the aligned posterior mean and $\sigma_j$ the standard deviation of latent coordinate $j$ over the training trajectories,
\begin{equation*}
    \text{NRMSE}_{\text{lat}} = \sqrt{\tfrac{1}{TBd}\textstyle\sum_{t,b,j} \big((\tilde{\mathbf{z}}_{t,b,j} - \mathbf{z}_{t,b,j}) / \sigma_j\big)^2}.
\end{equation*}
\paragraph{Dynamics NRMSE.} The learned drift is compared with the true drift at the same ground-truth test states: each true state $\mathbf{z}$ is mapped into the model's coordinates, $\mathbf{u} = (\mathbf{z} - \mathbf{b})\mathbf{A}^{-1}$, the learned field is evaluated there and mapped back, and
\begin{equation*}
    \text{NRMSE}_{\text{dyn}} = \frac{\sqrt{\tfrac{1}{TBd}\sum_{t,b} \|f_\phi(\mathbf{u}_{t,b}) A - f(\mathbf{z}_{t,b})\|^2}}{\sqrt{\tfrac{1}{T'B'd}\sum_{t,b} \|f(\mathbf{z}^{\text{train}}_{t,b})\|^2}},
\end{equation*}
so that the denominator is the RMS true drift on the training states. Because both fields are evaluated at the true states, latent-trajectory error does not enter this metric; it evaluates the analytic drift, not finite differences of noisy paths.
\paragraph{Forecasting NRMSE.} The second half of every test trajectory is hidden. Each method infers its posterior from the first half only, then propagates its learned prior SDE forward (128 sample paths) and we compare the mean forecast, mapped through the learned sensor, with the clean signal on the hidden half. The RMSE is divided by the pooled standard deviation of the per-channel-centered clean test signal, so that predicting the channel means scores 1.

\subsubsection{Kato Dataset}
There is no ground-truth latent, clean signal or drift, so the scores are observation-space and behavioral.
\paragraph{Reconstruction $R^2$.} $R^2$ between the reconstruction $\mathbf{C}_\phi\bar{\mathbf{z}}_t + \mathbf{d}_\phi$ and the observed standardized fluorescence, pooled over neurons.
\paragraph{Forecasting NRMSE.} The second half (100 frames, about 34\,s) of every 200-frame window is hidden, the posterior is inferred from the first half, and the learned prior is propagated across the hidden half (128 samples). The RMSE of the mean forecast against the observed frames is divided by the RMSE of persistence, i.e., of repeating the last observed frame, so values below 1 beat the persistence floor of that worm.
\paragraph{Behavioral decoding.} After the dynamical model is frozen we fit a class-balanced multinomial logistic regression from the posterior-mean latents to the annotated behavioral states. Features are standardized with training statistics; the inverse regularization strength $C \in \{10^{-3}, \dots, 10^{2}\}$ is selected by validation balanced accuracy (ties favor stronger regularization), and we report balanced accuracy (mean per-class recall) on held-out frames. Chance is $1/7$ for NoStim and $1/4$ for Stim worms.

\subsection{Additional Results}
\label{app:additional_results}
Table~\ref{tab:sim_results} summarizes the simulated benchmarks behind Fig.~\ref{fig:dynamics}D for ZENO, gpSLDS and SDE Matching. Six seeds were run per dataset and method; as in the figure, the seed with the worst dynamics NRMSE for each (dataset, method) pair is excluded and the remaining five are reported (this drops one diverged gpSLDS fit on Van der Pol, seed 4, whose reconstruction $R^2$ was $-2.85$; for the other pairs the dropped seed is an ordinary fit). ZENO reports its causal filtering posterior, whereas gpSLDS and SDE Matching report smoothing posteriors that condition on the full trajectory; the forecasting column is the one metric that is prefix-only for every method.

\begin{table}[htbp]
\centering
\caption{Simulated benchmarks: mean $\pm$ SD over 5 seeds. Reconstruction $R^2$ and RMSE are against the noiseless signal in standardized units; latent, dynamics and forecasting errors are normalized as defined in Section~\ref{app:metrics} (lower is better for all but $R^2$). ZENO scores are filtering estimates; the baselines are smoothers.}
\label{tab:sim_results}
\footnotesize
\setlength{\tabcolsep}{3pt}
\fitwidth{\begin{tabular}{llccccc}
\toprule
 & & \multicolumn{2}{c}{Reconstruction} & \multicolumn{3}{c}{NRMSE $\downarrow$} \\
\cmidrule(lr){3-4} \cmidrule(lr){5-7}
Dataset & Method & $R^2$ $\uparrow$ & RMSE $\downarrow$ & Latent & Dynamics & Forecast \\
\midrule
Double well & ZENO & $0.988 \pm 0.004$ & $0.027 \pm 0.003$ & $0.107 \pm 0.013$ & $0.342 \pm 0.085$ & $0.644 \pm 0.163$ \\
 & gpSLDS & $0.992 \pm 0.002$ & $0.023 \pm 0.003$ & $0.090 \pm 0.010$ & $0.321 \pm 0.188$ & $0.574 \pm 0.127$ \\
 & SDE Matching & $0.984 \pm 0.008$ & $0.030 \pm 0.005$ & $0.106 \pm 0.024$ & $0.488 \pm 0.201$ & $0.605 \pm 0.141$ \\
\midrule
Van der Pol & ZENO & $0.978 \pm 0.004$ & $0.048 \pm 0.002$ & $0.154 \pm 0.006$ & $0.271 \pm 0.030$ & $0.655 \pm 0.067$ \\
 & gpSLDS & $0.987 \pm 0.002$ & $0.038 \pm 0.002$ & $0.127 \pm 0.005$ & $0.192 \pm 0.047$ & $0.766 \pm 0.026$ \\
 & SDE Matching & $0.981 \pm 0.001$ & $0.045 \pm 0.001$ & $0.139 \pm 0.005$ & $0.251 \pm 0.054$ & $0.766 \pm 0.074$ \\
\midrule
Lorenz & ZENO & $1.000 \pm 0.000$ & $0.018 \pm 0.001$ & $0.014 \pm 0.001$ & $0.057 \pm 0.002$ & $0.029 \pm 0.002$ \\
 & gpSLDS & $1.000 \pm 0.000$ & $0.012 \pm 0.000$ & $0.009 \pm 0.000$ & $0.038 \pm 0.002$ & $0.031 \pm 0.001$ \\
 & SDE Matching & $1.000 \pm 0.000$ & $0.019 \pm 0.001$ & $0.010 \pm 0.001$ & $0.055 \pm 0.003$ & $0.065 \pm 0.007$ \\
\bottomrule
\end{tabular}}
\end{table}

Table~\ref{tab:kato_results} lists the per-worm scores behind Fig.~\ref{fig:kato}F--H. All scores follow the whole-trace protocol and are in-sample except the decoding probe, which holds out its own frames. gpSLDS forecast rollouts diverged on three animals (ratio to persistence above 10; marked ``div.''). Over the 12 animals, mean reconstruction $R^2$ is 0.830 (ZENO), 0.857 (gpSLDS) and 0.858 (SDE Matching); mean decoding balanced accuracy is 0.624, 0.524 and 0.482; and the mean forecast ratio is 0.78, 1.95 (over the nine non-diverged gpSLDS rollouts) and 0.71.
\begin{figure}
    \centering
    \includegraphics[width=1.0\linewidth]{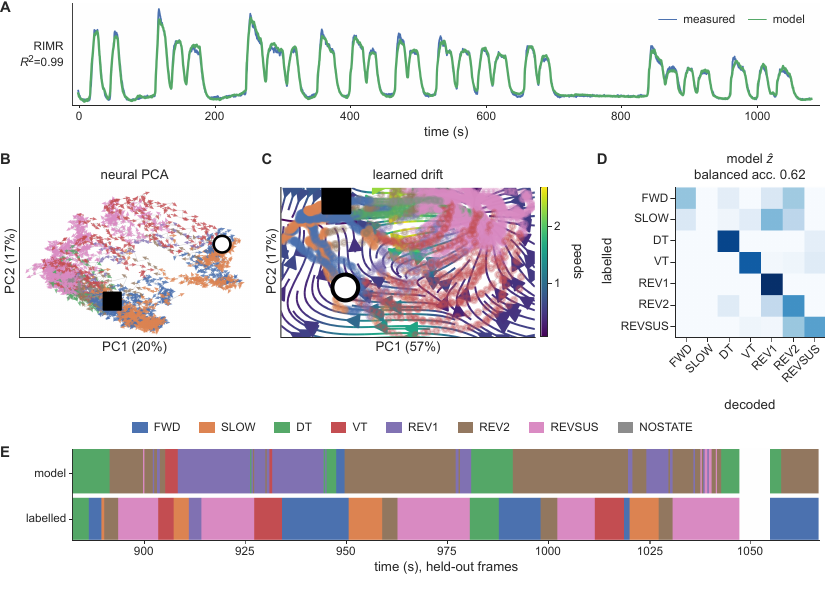}
    \caption{Example results for a NoStim worm from the Kato Dataset. See Fig \ref{fig:kato}A-E}
    \label{fig:No stim}
\end{figure}
\begin{table}[htbp]
\centering
\caption{Kato et al.\ (2015) per-worm results. $N$ is the number of recorded neurons and $T$ the recording length in seconds. For each method: reconstruction $R^2$ ($\uparrow$), behavioral decoding balanced accuracy ($\uparrow$; chance $1/7$ NoStim, $1/4$ Stim) and forecast RMSE relative to persistence ($\downarrow$; below 1 beats persistence). ``div.'' marks a diverged forecast rollout.}
\label{tab:kato_results}
\footnotesize
\setlength{\tabcolsep}{3pt}
\fitwidth{\begin{tabular}{lcc ccc ccc ccc}
\toprule
 & & & \multicolumn{3}{c}{ZENO} & \multicolumn{3}{c}{gpSLDS} & \multicolumn{3}{c}{SDE Matching} \\
\cmidrule(lr){4-6} \cmidrule(lr){7-9} \cmidrule(lr){10-12}
Worm & $N$ & $T$ (s) & $R^2$ $\uparrow$ & Decode $\uparrow$ & Forecast $\downarrow$ & $R^2$ $\uparrow$ & Decode $\uparrow$ & Forecast $\downarrow$ & $R^2$ $\uparrow$ & Decode $\uparrow$ & Forecast $\downarrow$ \\
\midrule
NoStim 0 & 109 & 1080 & 0.876 & 0.617 & 0.613 & 0.891 & 0.403 & 0.658 & 0.883 & 0.232 & 0.652 \\
NoStim 1 & 135 & 1080 & 0.862 & 0.537 & 0.761 & 0.893 & 0.250 & 0.731 & 0.884 & 0.255 & 0.701 \\
NoStim 2 & 131 & 1080 & 0.801 & 0.577 & 0.782 & 0.840 & 0.364 & 0.770 & 0.833 & 0.481 & 0.774 \\
NoStim 3 & 125 & 1080 & 0.894 & 0.492 & 0.730 & 0.906 & 0.499 & 0.806 & 0.891 & 0.464 & 0.628 \\
NoStim 4 & 129 & 1080 & 0.850 & 0.641 & 0.838 & 0.856 & 0.726 & div. & 0.859 & 0.631 & 0.813 \\
\midrule
Stim 0 & 107 & 720 & 0.811 & 0.724 & 0.916 & 0.874 & 0.706 & 0.733 & 0.865 & 0.630 & 0.703 \\
Stim 1 & 122 & 720 & 0.856 & 0.729 & 0.786 & 0.849 & 0.398 & 0.717 & 0.859 & 0.542 & 0.586 \\
Stim 2 & 124 & 720 & 0.800 & 0.601 & 0.853 & 0.899 & 0.464 & div. & 0.887 & 0.458 & 0.550 \\
Stim 3 & 134 & 720 & 0.782 & 0.664 & 0.742 & 0.819 & 0.688 & div. & 0.820 & 0.607 & 0.667 \\
Stim 4 & 123 & 720 & 0.830 & 0.524 & 0.942 & 0.838 & 0.504 & 4.333 & 0.858 & 0.436 & 1.090 \\
Stim 5 & 151 & 720 & 0.795 & 0.706 & 0.749 & 0.819 & 0.805 & 6.834 & 0.822 & 0.504 & 0.665 \\
Stim 6 & 146 & 720 & 0.804 & 0.677 & 0.675 & 0.795 & 0.480 & 1.959 & 0.835 & 0.538 & 0.730 \\
\bottomrule
\end{tabular}}
\end{table}

\end{document}